\documentclass[conference]{IEEEtran}
\IEEEoverridecommandlockouts

\usepackage{cite}
\usepackage{amsmath,amssymb,amsfonts}
\usepackage{algorithmic}
\usepackage{graphicx}
\usepackage{textcomp}
\usepackage{multirow}
\usepackage{xcolor}
\usepackage{booktabs}
\usepackage[hidelinks]{hyperref}
\def\BibTeX{{\rm B\kern-.05em{\sc i\kern-.025em b}\kern-.08em
    T\kern-.1667em\lower.7ex\hbox{E}\kern-.125emX}}
    
\makeatletter
\newcommand{\linebreakand}{%
  \end{@IEEEauthorhalign}
  \hfill\mbox{}\par
  \mbox{}\hfill\begin{@IEEEauthorhalign}
}
\makeatother

\begin{document}

\title{K-Gen: A Multimodal Language-Conditioned Approach for Interpretable Keypoint-Guided Trajectory Generation\\

\thanks{This research is funded by the Chongqing Natural Science Foundation Innovation and Development Joint Fund (Changan Automobile) (Grant No. CSTB2024NSCQ-LZX0157)}
}


\author{
  \IEEEauthorblockN{1\textsuperscript{st} Mingxuan Mu}
  \IEEEauthorblockA{\textit{Harbin Institute of Technology} \\
  \textit{Chongqing Research Inst. of HIT}\\
  mmxuan0516@163.com}
  \and
  \IEEEauthorblockN{2\textsuperscript{nd} Guo Yang}
  \IEEEauthorblockA{\textit{Changan Automobile Co., Ltd} \\
  Chongqing, China \\
  yangguo@changan.com.cn}
  \and
  \IEEEauthorblockN{3\textsuperscript{rd} Lei Chen}
  \IEEEauthorblockA{\textit{Changan Automobile Co., Ltd} \\
  Chongqing, China \\
  chenlei5@changan.com.cn}
  \linebreakand
  \IEEEauthorblockN{4\textsuperscript{th} Ping Wu}
  \IEEEauthorblockA{\textit{Changan Automobile Co., Ltd} \\
  Chongqing, China \\
  wuping@changan.com.cn}
  \and
  \IEEEauthorblockN{5\textsuperscript{th} Jianxun Cui*}
  \IEEEauthorblockA{\textit{Harbin Institute of Technology} \\
  \textit{Chongqing Research Inst. of HIT}\\
  cuijianxun@hit.edu.cn}
}

\maketitle

\begin{abstract}
Generating realistic and diverse trajectories is a critical challenge in autonomous driving simulation. While Large Language Models (LLMs) show promise, existing methods often rely on structured data like vectorized maps, which fail to capture the rich, unstructured visual context of a scene. To address this, we propose \textbf{K-Gen}, an interpretable keypoint-guided multimodal framework that leverages Multimodal Large Language Models (MLLMs) to unify rasterized BEV map inputs with textual scene descriptions. Instead of directly predicting full trajectories, K-Gen generates interpretable keypoints along with reasoning that reflects agent intentions, which are subsequently refined into accurate trajectories by a refinement module. To further enhance keypoint generation, we apply \textbf{T-DAPO}, a trajectory-aware reinforcement fine-tuning algorithm. Experiments on WOMD and nuPlan demonstrate that K-Gen outperforms existing baselines, highlighting the effectiveness of combining multimodal reasoning with keypoint-guided trajectory generation.
\end{abstract}


\section{Introduction}
Simulation is indispensable for autonomous driving research, enabling large-scale, risk-free testing\cite{fellendorf1994vissim,krajzewicz2002sumo,dosovitskiy2017carla}. Specifically, generating diverse, realistic traffic scenarios is crucial for evaluating motion planning algorithms. However, constructing high-fidelity simulation environments remains a challenging task: most existing approaches either rely on hand-crafted rules, which limit diversity, or data-driven generative models, which often struggle with controllability and interpretability.

\begin{figure}[htbp]
\centerline{\includegraphics[width=0.9\linewidth]{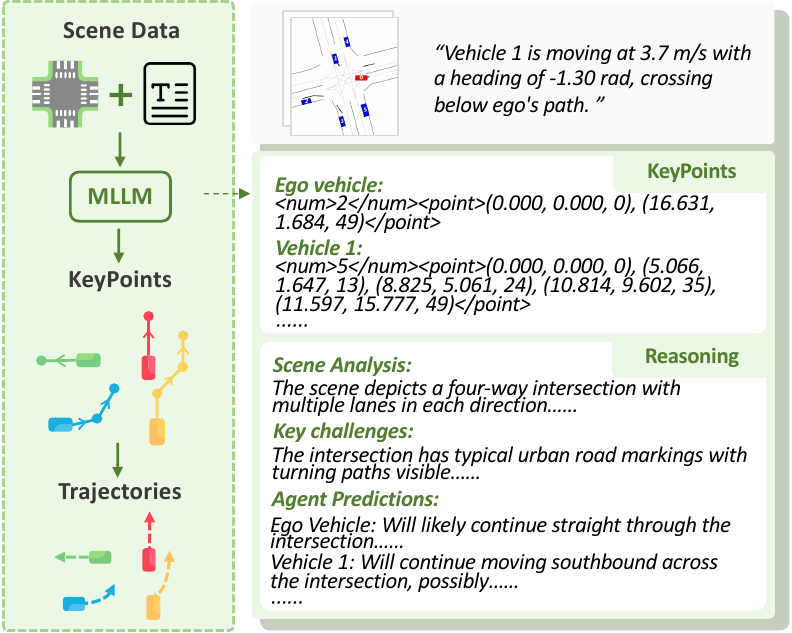}}
\caption{Overall workflow of the proposed multimodal trajectory generation framework K-Gen. The MLLM takes multimodal scene data (map images and textual inputs) as input, produces both reasoning and sparse keypoints, and refines them into complete trajectories.}
\vspace{-5mm}
\label{fig1}
\end{figure}

Recent advances in large language models (LLMs)\cite{achiam2023gpt,Anthropic2025Claude, zhu2025internvl3} have motivated a new paradigm of language-guided scenario generation, where textual reasoning can provide structured, human-readable explanations of agent behavior. While promising in terms of interpretability, these methods often suffer from coarse-grained motion control, limited physical consistency, and dependence on rigid intermediate representations that constrain generalization.

A critical limitation of prior works is their reliance on vectorized map inputs and structured agent representations, which fail to capture the full multimodal richness of driving environments. Vectorized encodings abstract away spatial details and contextual semantics essential for modeling complex interactions and reasoning about future motions. In contrast, raw visual representations of maps can preserve intricate lane structures, traffic elements, and local context, enabling more faithful and flexible understanding of the scene.

To overcome these challenges, we propose \textbf{K-Gen}, a multimodal large language model (MLLM) framework that combines visualized maps with textual scene inputs to generate accurate and interpretable trajectories. Instead of directly producing full trajectories, K-Gen first generates sparse keypoints guided by the Chain-of-Thought (CoT) reasoning of MLLM, which are then refined into complete trajectories by the TrajRefiner module. Extensive experiments on WOMD\cite{ettinger2021large} and nuPlan\cite{caesar2021nuplan} demonstrate that our method outperforms existing baselines in trajectory quality.

Our main contributions are threefold:
\begin{itemize}
\item We propose \textbf{K-Gen}, a multimodal trajectory generation framework that integrates rasterized maps and text inputs, enabling both interpretable intent prediction and accurate trajectory forecasting.
\item We design a \textbf{keypoint-guided strategy} that decomposes the trajectory generation task into two steps: a keypoint generation phase and a trajectory refinement phase. This design effectively improves accuracy and stability compared to direct MLLM outputs.
\item We introduce \textbf{T-DAPO}, a Trajectory-aware Decoupled Clip and Dynamic Sampling Policy Optimization (T-DAPO) algorithm that incorporates trajectory-centric reward signals to enhance keypoint generation and ensure accurate motion reconstruction.
\end{itemize}

\section{Related Work}

\subsection{Trajectory Generation}
Trajectory generation aims to synthesize realistic agent motions in complex scenarios. Early model-based methods\cite{dosovitskiy2017carla,wheeler2016factor} rely on predefined rules, while recent learning-based approaches\cite{zhong2022guided,shi2022motion,wu2024smart} learn distributions directly from data. However, relying on structured inputs without capturing high-level semantics severely limits their flexibility and controllability.

Recently, language-guided trajectory generation has emerged as a promising direction. LCTGen\cite{tan2023language} pioneers this direction by converting language queries to structured representations with LLMs, which are then decoded into realistic traffic scenarios. Building on this, InteractTraj\cite{xia2024language} introduces interaction-aware codes to enhance inter-agent dynamics in generated trajectories. Traj-LLM\cite{yang2025trajectory} takes a different approach by directly producing trajectories through LLMs with its "interaction-behavior-trajectory" translation framework. LC-LLM\cite{peng2025lc} further leverages chain-of-thought reasoning for trajectory prediction, generating both trajectory forecasts and lane-change intents in natural language.

Despite these advances, most existing methods still rely on vectorized maps or structured representations, which fail to exploit richer multimodal cues from realistic traffic scenes. Our K-Gen framework adopts a multimodal design, where maps are provided in visualized image form and combined with textual reasoning. This integration enables more expressive modeling of interactions and intentions for trajectory generation.

\subsection{Reinforcement Fine-Tuning}
Reinforcement Fine-Tuning (RFT) has evolved from general policy gradient methods\cite{williams1992simple,schulman2017proximal} to specialized algorithms for language model alignment. Early approaches established foundational optimization techniques\cite{schulman2015trust}, while recent advances focus on direct preference optimization through methods like DPO\cite{rafailov2023direct} that eliminate explicit reward modeling. Subsequent developments introduced group-relative optimization strategies, with GRPO\cite{shao2024deepseekmath} incorporating gradient regularization and DAPO\cite{yu2025dapo} further enhancing training stability through differentiable alignment objectives. These methods collectively address key challenges in preference-based reinforcement learning, including training stability, reward modeling, and policy alignment.

In autonomous driving, AlphaDrive\cite{jiang2025alphadrive} employs autonomous driving planning with GRPO and a two-stage reasoning framework. AutoVLA\cite{zhou2025autovla} advances end-to-end driving by combining vision-language-action pretraining with GRPO-based fine-tuning. 
Our approach builds upon DAPO's stable alignment framework, incorporating chain-of-thought reasoning and enhanced optimization strategies for trajectory generation.

\begin{figure*}[htbp]
\centerline{\includegraphics[width=1.0\linewidth]{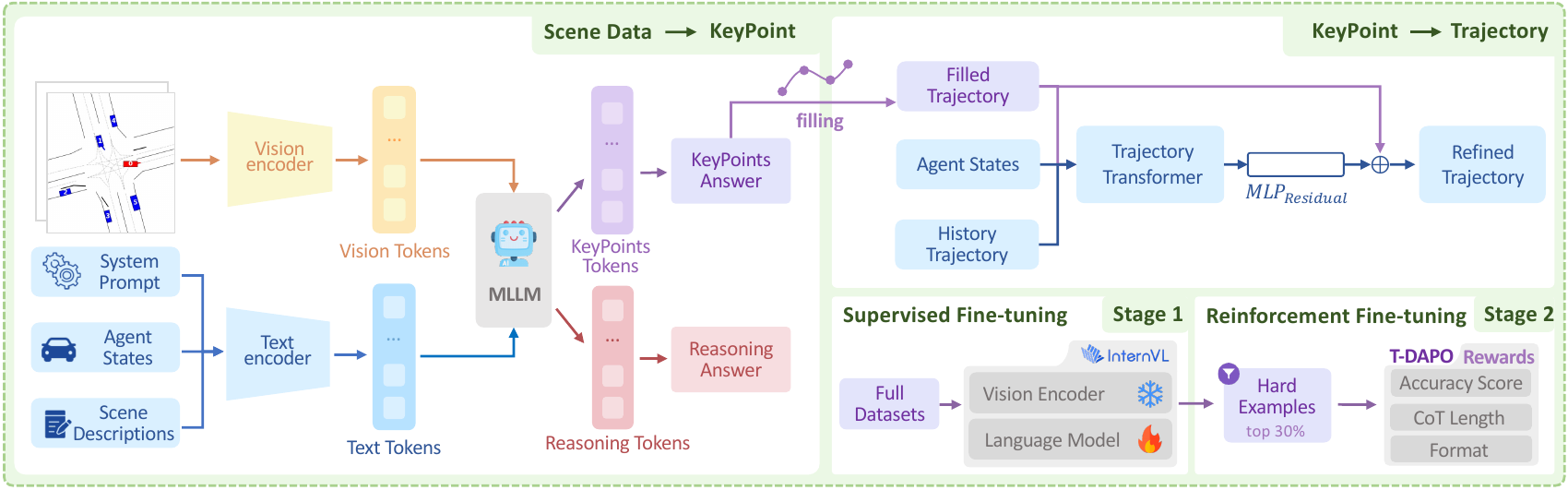}}
\caption{Overall framework of K-Gen. The framework is composed of two main components: an MLLM for reasoning and keypoint generation, and a Transformer-based module for trajectory refinement. For the MLLM, the training pipeline includes two stages: SFT with CoT outputs, followed by reinforcement RFT using the T-DAPO algorithm.}
\vspace{-3mm}
\label{fig2}
\end{figure*}

\section{Method}

\subsection{Overview}
Fig.~\ref{fig2} illustates the two parts of our K-GEN: first generating sparse keypoints guided by multimodal reasoning, and then refining them into full-length trajectories. 

At the keypoint generation module, a vision encoder and a text encoder extract tokens from the BEV map and text inputs, which are integrated through a MLLM. The MLLM produces both CoT reasoning and structured keypoint sequences, effectively capturing agents’ high-level intentions. At the trajectory refinement module, the interpolated trajectories derived from the keypoints are further corrected by the TrajRefiner. This module incorporates historical trajectories and agent states to predict residual adjustments, ensuring the final outputs are both smooth and accurate. 

For training, we propose the \textbf{T}rajectory-aware \textbf{D}ecoupled Clip and \textbf{D}ynamic s\textbf{A}mpling \textbf{P}olicy \textbf{O}ptimization (T-DAPO) algorithm, a trajectory-refined variant of the DAPO algorithm. By emphasizing challenging samples and rewarding high-quality outputs, T-DAPO enables our model to generate interpretable and high-fidelity trajectories in complex traffic scenarios.

\subsection{Data Preprocessing Pipeline}
We design a multi-stage preprocessing pipeline to transform raw trajectory data into structured representations, generate reasoning annotations, and extract motion-critical keypoints, compressing temporal redundancy while preserving kinematic features.

\textbf{Scene Description.} For each scenario $\tau_i$, the $i^{th}$ agent is represented as a tuple $A_{i}=\left(\text {type}_{i}, p_{i}^{0}, v_{i}^{0}, \Delta d_{i}, \Delta \theta_{i}\right)$, where $\text {type}_{i}$ denotes the agent type(vehicle, pedestrian, cyclist), $p_{i}^{0}$ and $v_{i}^{0}$ are initial position and velocity, and $\left(\Delta d_{i}, \Delta \theta_{i}\right)$ denote relative distance and bearing to the ego agent. These tuples are serialized into natural language to provide the MLLM with a structured scene context.

\textbf{Reasoning Data.} We generate structured reasoning annotations by prompting Claude 3.7 Sonnet\cite{Anthropic2025Claude} to analyze traffic scenes through three interconnected perspectives: road geometry, potential collision risks, and intention prediction. The resulting annotations serve as auxiliary supervision to instill structured reasoning capabilities prior to trajectory generation.

\textbf{KeyPoints.}We extract spatial keypoints $K_g = \text{DP}(\tau_i(t))$ via the Douglas-Peucker algorithm\cite{douglas1973algorithms} for high-curvature segments, and kinematic keypoints $K_v = \{p_{i}^{t} \mid |\Delta v_i^{t}| > \delta_v\}$ for significant velocity changes. The final keypoint set combines both aspects through the union operation $K = K_g \cup K_v$, ensuring a comprehensive representation that captures both the geometric shape and dynamic motion characteristics of trajectories.

\subsection{SFT-based Warm-up}
In the warm-up stage, we adapt the pretrained MLLM to the trajectory generation task through supervised fine-tuning. The model’s output consists of two distinct token types: reasoning tokens $y_R$ and keypoint tokens $y_K$, which together form the complete output sequence $y^*_t=(y_1^*, \ldots, y_R^*,y_{R+1}^*, \ldots,y_{R+K}^*)$, where $T=R+K$ is the total sequence length. Given an input $x$ comprising multimodal scene information such as map image, scene description, contextual instructions, and historical trajectories, the model is trained to maximize the conditional likelihood of the ground-truth output sequence:
\begin{equation}
\mathcal{L}_{\text{SFT}}=-\sum_{t=1}^{T}\log P_{\theta}\left(y_{t}^{*}\mid x, y_{<t}^{*}\right),\quad T=R+K\label{eq1}
\end{equation}

\subsection{RFT-based Enhancement}
To further enhance the model beyond the SFT warm-up, we employ the proposed Trajectory-aware Decoupled Clip and Dynamic Sampling Policy Optimization (T-DAPO) algorithm. T-DAPO is specifically designed for the trajectory generation task, where model performance critically depends on both the fidelity of predicted trajectories and the interpretability of reasoning. Instead of treating all samples equally, T-DAPO emphasizes the most challenging subset of the training data to encourage the model to improve in difficult scenarios. Specifically, we define this subset as the top 30\% of samples exhibiting the highest mean Average Displacement Error (mADE) and mean Final Displacement Error (mFDE) based on the predictions of the post-SFT model. Continuous trajectory spaces cause severe gradient oscillation in standard GRPO. T-DAPO's truncation mechanism stabilizes training on these hard samples, preventing convergence to simplistic straight-line predictions.

\textbf{Accuracy Reward. }This component evaluates the predicted trajectory quality by computing trajectory metrics via the TrajRefiner module, rewarding lower errors.

\begin{equation}
R_{\text{acc}} = \lambda_1 \cdot e^{-\text{ADE}/\tau_1} + \lambda_2 \cdot e^{-\text{FDE}/\tau_2}\label{eq2}
\end{equation}
where $\tau_1$ and $\tau_2$ serve as normalization coefficients that control the decay scale of trajectory errors, and $\lambda_1$ and $\lambda_2$ are scaling factors balancing the ADE and FDE terms.

\textbf{CoT Length Reward.} This component encourages concise yet informative reasoning by penalizing unnecessarily long Chain-of-Thought outputs: 
\begin{equation}
R_{\text{cot}} =
\begin{cases}
1-\frac{L}{L_{\max}}, & \mathrm{if~}L\leq L_{\max}, \\
0, & \mathrm{if~}L>L_{\max}, 
\end{cases}\label{eq3}
\end{equation}
where \(L\) denotes the actual length, and 
$L_{\max}$ represents the maximum allowable length.

\textbf{Format Correctness Reward.} This component ensures structural integrity by verifying the presence and ordering of required tags (e.g., \texttt{<think>}, \texttt{<answer>}, \texttt{<point>}, \texttt{<num>}): 
\begin{equation}
R_{\text{fmt}}=
\begin{cases}
1, & \text{if all tags are present and correctly ordered,} \\
0, & \text{otherwise.} 
\end{cases}\label{eq4}
\end{equation}

The final composite reward is obtained as a weighted sum:
\begin{equation}
R = \alpha R_{\text{acc}} + \beta R_{\text{cot}} + \gamma R_{\text{fmt}}\label{eq5}
\end{equation}
where $\alpha, \beta$, and $\gamma$ are hyperparameters weighting the relative importance of accuracy, reasoning conciseness, and format correctness, respectively.

\subsection{TrajRefiner}
We design TrajRefiner as a Transformer-based residual refinement module to adjust filled trajectories derived from sparse keypoints. Given historical trajectories $H$, sparse keypoints $K$, and initial agent states $S$, we first obtain a coarse filled trajectory $\tilde{Y}$ via linear filling. TrajRefiner then predicts a residual correction $\Delta Y$ to refine $\tilde{Y}$ into the final trajectory $\hat{Y} = \tilde{Y} + \Delta Y$. 

The model architecture employs two Transformer decoder layers\cite{xia2024language}, where $K$ interacts with both $H$ and $S$ through cross-attention. The resulting feature streams are fused via another cross-attention block, enabling joint reasoning over temporal dynamics and static state cues. An MLP head $MLP_{Residual}$ finally outputs the residual correction $\Delta Y$.

The overall training objective integrates three key components: a motion loss $\mathcal{L}_{\text{motion}}$ for trajectory accuracy, a kinematic consistency loss $\mathcal{L}_{\text{KCL}}$ to enforce feasibility, and a final point loss $\mathcal{L}_{\text{FPL}}$ to ensure endpoint precision. The orientation $\theta$ and velocity $v$ used in $\mathcal{L}_{\text{KCL}}$ are derived from the predicted trajectories. Each term is implemented using mean squared error (MSE), collectively ensuring that the refined trajectories are both accurate and kinematically consistent:

\vspace{-1.5mm}
\begin{gather}
\mathcal{L}_{\mathrm{motion}}=\mathrm{MSE}(\hat{p},p^*) \nonumber , \text{ }  \mathcal{L}_{\mathrm{FPL}}=\mathrm{MSE}(\hat{p}_T,p_T^*) \nonumber \\
\mathcal{L}_{\mathrm{KCL}}=\lambda_\theta\mathrm{MSE}(\hat{\theta},\theta^*)+\lambda_v\mathrm{MSE}(\hat{v},v^*)  \\
\mathcal{L}=\mathcal{L}_\mathrm{motion}+\mathcal{L}_\mathrm{KCL}+\mathcal{L}_\mathrm{FPL} \nonumber
\end{gather}

\begin{table}[ht]
    \centering
    \small
    \vspace{-3mm}
    \caption{\centering \textsc{Quantitative results on WOMD and nuPlan datasets. Methods marked with $\dagger$ are based on closed-source LLMs (API access only, no model training).}}
    \scalebox{0.95}{
    \begin{tabular}{l c c c c c}
    \toprule
         Dataset & Method & Size & mADE$\downarrow$ & mFDE$\downarrow$ & SCR$\downarrow$ \\
        \midrule
        \multirow{8}{*}{WOMD}
         & LCTGen\cite{tan2023language}$^\dagger$ & - & 1.262& 2.696& 0.072\\
         & InteractTraj\cite{xia2024language}$^\dagger$ & - & 1.067& \textbf{2.190}& 0.070\\
         & InternVL3\cite{zhu2025internvl3}& 2B & 1.260 & 3.147 & 0.011\\
         & InternVL2.5\cite{chen2024expanding}& 4B & 1.187 & 2.796 & 0.010\\
         & InternVL3\cite{zhu2025internvl3}& 8B & 1.128 & 2.770 & 0.007\\
         & Qwen2.5-VL\cite{Qwen2.5-VL}& 7B & 0.989 & 2.602 & 0.008\\
         & Qwen3-VL\cite{Qwen3-VL}& 8B & 0.926 & 2.478 & 0.007\\
         & \textbf{K-Gen (ours)} & 8B & \textbf{0.915} & 2.422 & \textbf{0.006}\\
        \midrule
        \multirow{8}{*}{nuPlan}
         & LCTGen\cite{tan2023language}$^\dagger$ & - & 1.161& 2.497& 0.074\\
         & InteractTraj\cite{xia2024language}$^\dagger$ & - & 0.962& 1.987& 0.067\\
         & InternVL3\cite{zhu2025internvl3}& 2B & 0.676 & 1.654 & 0.028\\
         & InternVL2.5\cite{chen2024expanding}& 4B & 0.666 & 1.669 & 0.029\\
         & InternVL3\cite{zhu2025internvl3}& 8B & 0.610 & 1.514 & 0.029\\
         & Qwen2.5-VL\cite{Qwen2.5-VL}& 7B & 0.624 & 1.547 & 0.028\\
         & Qwen3-VL\cite{Qwen3-VL}& 8B & 0.607 & 1.470 & 0.028\\
         & \textbf{K-Gen (ours)} & 8B & \textbf{0.591} & \textbf{1.478} & \textbf{0.027}\\
        \bottomrule
        \label{tab:main_results}
    \end{tabular}}
    \vspace{-0.5cm}
\end{table}

\begin{figure*}[htbp]
\centerline{\includegraphics[width=0.9\linewidth]{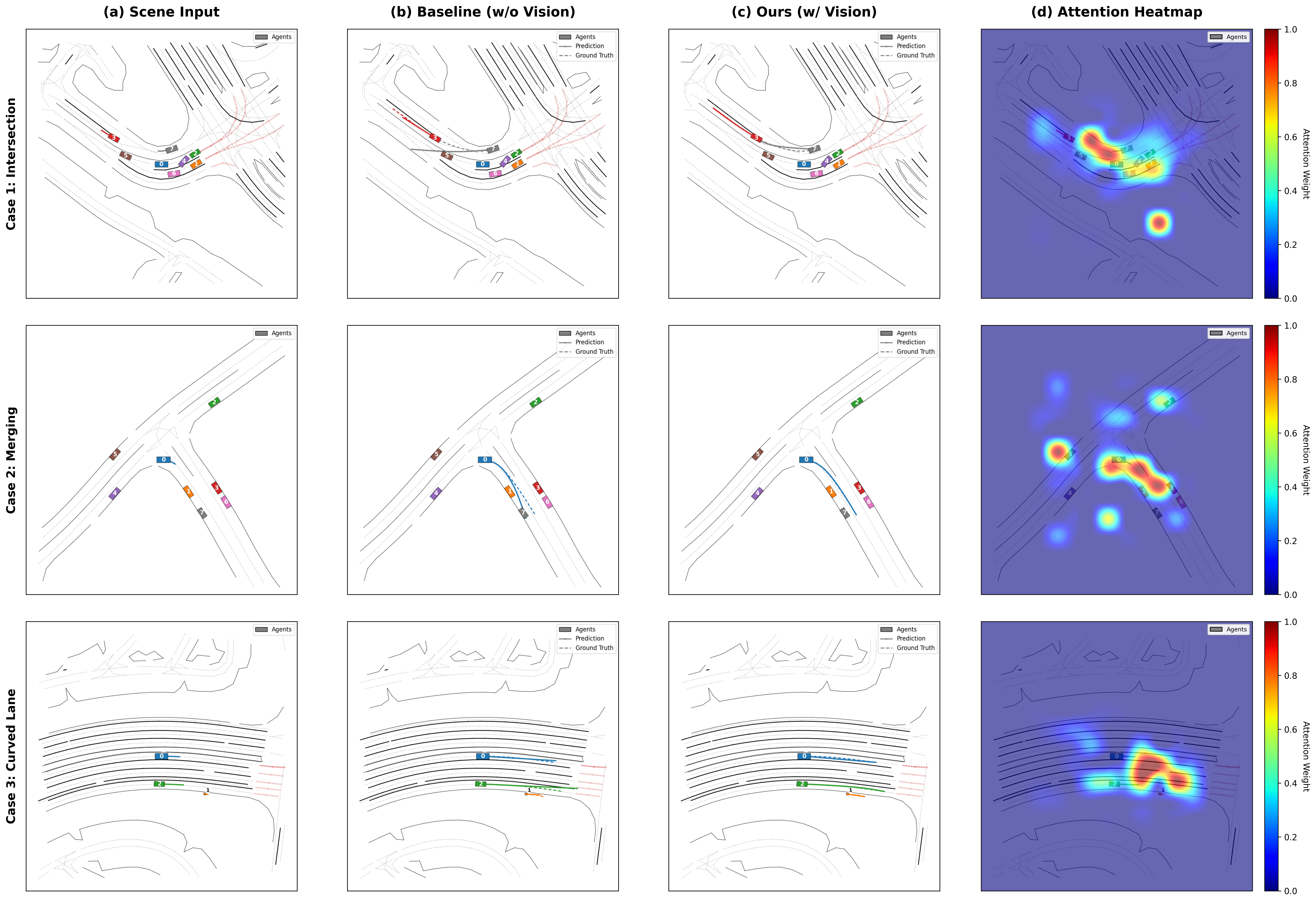}}
\caption{Qualitative results and attention visualization across diverse scenarios (Case 1: Intersection, Case 2: Merging, Case 3: Curved Lane). Columns (a)-(c) show the comparison between the ground truth and trajectories generated by the baseline and our TrajRefiner. Column (d) visualizes the MLLM's attention heatmaps, where warmer colors indicate higher focus on safety-critical regions and interacting agents.}
\label{fig3}
\end{figure*}

\begin{table*}[t]
\centering
\caption{\centering \textsc{Ablation study on training pipeline components. All experiments are based on the 8B model and WOMD datasets.}}
\begin{tabular}{l c c c c  |c cc}
\toprule
Setting & SFT &  GRPO&DAPO & TrajRefiner & mADE $\downarrow$ &mFDE$\downarrow$ &SCR$\downarrow$\\
\midrule
Zero-shot& - & - & - & - & 9.880 & 15.369 & 0.5034\\
SFT& \checkmark & -& - & - & 1.128 & 2.770 & 0.0068\\
SFT + TrajRefiner& \checkmark & - & - & \checkmark & 0.926 & 2.496 & 0.0073\\
SFT + GRPO +  TrajRefiner& \checkmark & \checkmark & - & \checkmark &   0.923 & 2.447 & 0.0064\\
\midrule
SFT + DAPO + TrajRefiner(ours)& \checkmark &  -&\checkmark & \checkmark &   \textbf{0.915}& \textbf{2.422}&\textbf{0.0058}\\
\bottomrule
\label{tab:pipeline_ablation}
\end{tabular}
\end{table*}

\section{Experiments}

\subsection{Experimental Setup}
\textbf{Datasets.} We evaluate our method on two datasets, Waymo Open Motion Dataset (WOMD)\cite{ettinger2021large} and nuPlan\cite{caesar2021nuplan}. Both datasets provide diverse urban driving scenarios with multi-agent interactions. To balance inference speed and accommodate the context length limits of current MLLMs, we set the maximum number of agents per scenario to 8, sorted by their entering time. All evaluation metrics are calculated exclusively based on these filtered agents to ensure fair comparisons. We adopt a 5-second horizon at 10 fps, resulting in 50 timesteps each scene.

\textbf{Metrics.} For trajectory evaluation, we report the mean average displacement error (mADE), the mean final displacement error (mFDE), and the scenario collision rate (SCR) to measure accuracy and safety\cite{tan2023language}.

\textbf{Implementation.} We use InternVL3-8B\cite{zhu2025internvl3} as the base model. For keypoint extraction, we set the simplification threshold $\epsilon=0.5$ and the velocity change threshold $\delta v=1.0$. In SFT stage, reasoning data is generated with Claude 3.7 Sonnet, and the model is full-parameter fine-tuned with a learning rate of $4\mathrm{e}{-5}$. The TrajRefiner module is trained separately with AdamW (learning rate $3\mathrm{e}{-4}$, batch size 64, 100 epochs), using loss weights $\lambda_{\theta}=1.0$ and $\lambda_{v}=0.1$. For the T-DAPO algorithm, we set the maximum CoT length $L_{\max}$ to 512. The reward scaling factors are set to $\lambda_1=\text{0.5}, \lambda_2=\text{0.5}, \tau_1=\text{1.0}$, and $\tau_2=\text{2.0}$, with the final composite reward weights configured as $\alpha=\text{0.7}, \beta=\text{0.2}$, and $\gamma=\text{0.1}$. These hyperparameters were empirically tuned to balance error scales, prioritizing safety and physical compliance. All experiments are run on eight NVIDIA A100 GPUs. For inference efficiency, benchmarking 200 scenarios on a single A100 yields an average running time of 1.63 seconds per scene, confirming practical feasibility.

\subsection{Main Results}
Table~\ref{tab:main_results} reports the quantitative results on the WOMD and nuPlan datasets. We compare our method with LCTGen~\cite{tan2023language} and InteractTraj~\cite{xia2024language}. While InteractTraj achieves a lower mFDE on the WOMD dataset, K-Gen significantly outperforms baselines in mADE and SCR. This confirms that our approach prioritizes interaction-aware, physically consistent trajectories and safety over aggressively overfitting the endpoint accuracy. We also evaluate InternVL models of different scales (2B, 4B, and 8B) under the SFT+interpolation setting. While larger models show gradual improvements, our K-Gen significantly outperforms them. We also evaluated Qwen2.5-VL and Qwen3-VL. K-Gen outperforms them, as their dynamic image cropping introduces coordinate inconsistencies that weaken fine-grained spatial modeling. This highlights that the proposed training strategy and TrajRefiner module brings substantial gains beyond simple model scaling.

\subsection{Qualitative Analysis}
To further investigate how our model perceives complex driving environments, we visualize the attention heatmaps of the MLLM backbone across diverse scenarios, as shown in Fig.~\ref{fig3}.

\textbf{Visualizing Attention Patterns.} 
The attention heatmaps (column d) reveal that our model effectively focuses on the most safety-critical regions relevant to the ego-vehicle's future path. 
\begin{itemize}
    \item \textbf{Intersection (Case 1):} When navigating a busy intersection, the model's attention is highly concentrated on the cluster of interacting agents and the upcoming junction area, ensuring a balanced consideration of multi-agent dynamics.
    \item \textbf{Merging (Case 2):} In the merging scenario, the attention peak accurately aligns with the conflict point where the merging vehicle interacts with the main road traffic, demonstrating the model's ability to identify potential collision risks.
    \item \textbf{Curved Lane (Case 3):} For the curved lane scenario, the heatmap shows a continuous distribution along the lane boundaries and the leading vehicles, indicating that the model successfully captures the geometric constraints of the road.
\end{itemize}

These qualitative results suggest that the MLLM does not merely process the scene as a holistic image but actively "reasons" about the spatial relationships and key constraints, which explains the superior trajectory accuracy and safety observed in the quantitative experiments.

\subsection{Ablation Study}

To validate our design, we perform ablation studies on both the training pipeline and the TrajRefiner module, disentangling the contributions of each component.

\textbf{Effects of the training pipeline components.} The ablation study evaluates the impact of key training pipeline components: supervised fine-tuning, reinforcement learning with GRPO and DAPO, and the TrajRefiner module. Our results, summarized in Table \ref{tab:pipeline_ablation}, demonstrate that each stage contributes to significant performance gains on the WOMD datasets. Specifically, SFT provides a strong foundation, the TrajRefiner module enhances trajectory accuracy, and the reinforcement learning stages further refine the results. The results confirm the effectiveness of each component and highlight the robustness of our overall framework.

\textbf{Effects of TrajRefiner.} We evaluate the contributions of relative coordinate encoding (RCE), kinematic consistency loss (KCL), and final point loss (FPL) within the TrajRefiner module. As shown in Table \ref{tab:TrajRefiner_ablation}, the linear interpolation baseline yields the highest error. Incorporating RCE, which transforms the trajectory into an agent-centric relative coordinate system, leads to substantial improvements. Building upon this, both KCL and FPL further refine the trajectory and enhance the evaluation metrics. The full integration of all three components in our complete model, TrajRefiner, proves to be the most effective, achieving the best overall performance. Crucially, TrajRefiner corrects physically infeasible keypoints from the MLLM—even when textual reasoning is correct—using residual predictions and kinematic consistency loss. The near-zero SCR validates its capability.

\begin{table}[h]
\centering
\caption{\centering \textsc{Ablation study of TrajRefiner. Experiments are based on the WOMD datasets.}}
\begin{tabular}{lccc}
\toprule
Method & mADE$\downarrow$& mFDE$\downarrow$& SCR$\downarrow$\\
\midrule
Baseline& 1.154 & 2.660 & 0.0229\\
+ RCE & 0.499 & 1.404 & 0.0116 \\
+ KCL & 0.416 & 1.289 & 0.0099 \\
+ FPL & 0.423 & 0.798 & 0.0086 \\
\midrule
TrajRefiner(ours)& \textbf{0.369} & \textbf{0.738} & \textbf{0.0078} \\
\bottomrule
\label{tab:TrajRefiner_ablation}
\end{tabular}
\vspace{-0.5cm}
\end{table}

\section{Conclusion}
We introduced K-Gen, a multimodal framework for interpretable, high-fidelity trajectory generation. By integrating rasterized map inputs and separating strategic intent from motion execution, K-Gen provides a new paradigm for language-guided modeling. Extensive experiments on WOMD and nuPlan validate its superior accuracy, safety, and generalization in diverse autonomous driving scenarios.

\bibliographystyle{IEEEtran}
\bibliography{refs}

\end{document}